\documentclass{article}
\usepackage{ijcai23}

\usepackage{times}
\usepackage{soul}
\usepackage{url}
\usepackage[hidelinks]{hyperref}
\usepackage[utf8]{inputenc}
\usepackage[small]{caption}
\usepackage{graphicx,xcolor}
\usepackage{amsmath}
\usepackage{amsthm}
\usepackage{booktabs}
\usepackage{algorithm}
\usepackage{algorithmic}
\usepackage[switch]{lineno}

\usepackage[inkscapelatex=false]{svg}
\usepackage{MnSymbol}
\usepackage{makecell}
\usepackage{wrapfig}
\usepackage{subcaption}
\usepackage[export]{adjustbox}

\title{Multivariate Time Series Anomaly Detection via Dynamic Graph Forecasting}

\author{
Katrina Chen
\and
Mingbin Feng\and
Tony S. Wirjanto
\affiliations
Department of Statistic and Actuarial Science, University Of Waterloo
\emails
j385chen@uwaterloo.ca,
ben.feng@uwaterloo.ca,
twirjanto@uwaterloo.ca
}

\begin{document}

\maketitle

\begin{abstract}
Anomalies in univariate time series often refer to abnormal values and deviations from the temporal patterns from majority of historical observations.
In multivariate time series, anomalies also refer to abnormal changes in the inter-series relationship, such as correlation, over time.
Existing studies have been able to model such inter-series relationships through graph neural networks.
However, most works settle on learning a static graph globally or within a context window to assist a time series forecasting task or a reconstruction task, whose objective is not tailored to explicitly detect the abnormal relationship. Some other works detect anomalies based on reconstructing or forecasting a list of inter-series graphs, which inadvertently weakens their power to capture temporal patterns within the data due to the discrete nature of graphs.
In this study, we propose DyGraphAD, a multivariate time series anomaly detection framework based upon a list of dynamic inter-series graphs. The core idea is to detect anomalies based on the deviation of inter-series relationships and intra-series temporal patterns from normal to anomalous states, by leveraging the evolving nature of the graphs in order to assist a graph forecasting task and a time series forecasting task simultaneously. Our numerical experiments on real-world datasets demonstrate that DyGraphAD has superior performance than baseline anomaly detection approaches.

\end{abstract}

\section{Introduction}
\label{sec: introduction}

%
In the current era of Industry 4.0, more and more critical infrastructures are controlled by cyber-physical systems. Due to the complexity of these systems, unanticipated events, such as hardware failures or cyber-attacks, can cause system downtime and may even lead to catastrophic failures. To combat adversarial scenarios, such systems rely on "internet of things" (IoT) technologies for real-time monitoring. Given the massive amount of time series data generated continuously from IoT sensors, there is a surge of interest in developing efficient anomaly detection techniques to automatically identify unexpected system behaviors, i.e., anomalies, and trigger an alarm quickly.

\begin{figure}[t]
\centering
\includegraphics[width=1\linewidth]{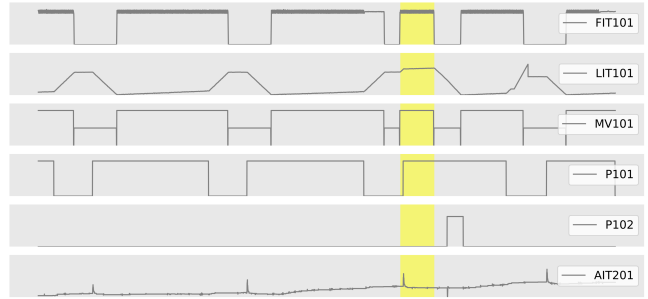}

\caption{An illustrative example (from SWaT dataset) showing that the abnormal event marked as yellow can be detected via inter-series relationship.}
\label{fig:swat_normal_abnormal}
\vspace{-0.1cm}
\end{figure}

For univariate time series, anomaly detection techniques generally rely on finding the points or sub-sequences that break the temporal dynamics, such as the abnormal change of periodicity or trend,  with respect to the past sequences. As the time series data collected from sensors are typically high-dimensional, detection techniques can leverage feature-wise information, in addition to temporal information within each time series, to identify anomalies. For example, Figure~\ref{fig:swat_normal_abnormal} shows a time series segment for which an industrial system operates in a normal state followed by an abnormal event. It is difficult to identify the anomaly highlighted in the yellow region based on purely irregular temporal patterns. However, it can be easily identified via the abnormal inter-series relationship, i.e., the value of ``LIT101'' stays high in the yellow region, which should be decreasing when the value of ``FIT101'' is high. In essence, learning both feature-wise relationship and temporal-wise dynamics is essential to improving the performance of multivariate time series anomaly detection.

With recent advances in neural networks and their ability to model various types of data in a task-specific way, deep learning approaches have been able to deliver promising results in the field of anomaly detection. The majority of deep time series anomaly detection methods employ recurrent neural networks (RNNs)~\cite{malhotra2016lstm}~\cite{hundman2018detecting} or temporal convolution neural networks (TCNs)~\cite{he2019temporal}~\cite{munir2018deepant} to model the temporal dependency within time series. However, RNNs and TCNs are not designed to explicitly learn the relationship among time series, thus limiting their ability to identify anomalies and provide interpretable results based on the deviation of inter-series relationship from normal to abnormal states.
 
Recent studies leverage the power of graph neural networks (GNNs)~\cite{zhou2020graph} to model the relationship among time series. However, unlike graph-structured data with readily available adjacency matrices, GNNs are not directly applicable to multivariate time series data, as the graph structure among time series is often not given in most applications. Thus, existing works either induce an inter-series graph structure through graph structure learning techniques~\cite{deng2021graph}~\cite{zhao2020multivariate}~~\cite{han2022learning}~\cite{han2022learning} or construct the graph based on similarity of pairwise time series before the training~\cite{hu2021time}~\cite{wu2021event2graph}~\cite{zhang2019deep}. In particular, GDN~\cite{deng2021graph} and MTAD-GAT~\cite{zhao2020multivariate} feed one-hot embeddings or time series segments as node features. However, as the graph structure is framed to be static either globally or within a context window in these studies, it is not tailored specifically to detect the abnormal inter-series relationship evolving between normal and anomalous states. In contrast, MSCRED~\cite{zhang2019deep} pre-computes a series of inter-series similarity matrices and detects anomalies through a graph reconstruction process. However, the graph reconstruction task is inadequate to capture finer temporal anomalies primarily due to the discrete nature of the graphs, i.e., graphs are apart from each other to capture a long time span, thus, coarsening the original time series.

To improve upon the above existing approaches, we propose a novel framework, DyGraphAD, that aims at detecting both the abnormal inter-series and intra-series patterns effectively. The framework is built upon a list of pre-constructed dynamically evolving similarity graphs aiming at incorporating the inter-series relationship changes to serve as an inductive bias. The graphs are used to assist two distinct forecasting tasks based on either historical graphs or past sequences, with each focusing on learning the feature-wise or temporal-wise patterns in the normal states, whereas anomalies are identified based on their combined forecasting errors.

We summarize our contributions as follows:
\begin{itemize}
    \item We propose a dynamic graph forecasting framework for multivariate time series anomaly detection. The core component of the framework is a graph encoder module built upon a series of dynamically evolving inter-series graphs, which is designed to model the short-term relationship variation as well as the long-term static relationship among time series.
    \item We couple a graph forecasting task with a time series forecasting task through the graph encoder module. Both tasks benefit from the evolving pattern of the dynamic graphs. More importantly, combining the error score of both tasks compensates for the drawbacks of one another and together improves the anomaly detection performance.
    \item We conduct a series of experiments on real-world industrial benchmark datasets. The experiment results and case study show that our model is superior in terms of detection accuracy compared to baseline methods.
\end{itemize}

\section{Related Works}
\subsection{Time Series Anomaly Detection}
Time series anomaly detection is the process of identifying the data point or short temporal sequence that rarely occurred in the past. In general, the training scheme of a time series anomaly detection method can be classified into supervised, unsupervised, or semi-supervised learning~\cite{chandola2009anomaly}. A supervised setting requires the existence of both abnormal and normal data instances, which is often an impractical approach due to the difficulties in collecting large-scale labeled data (especially for rare abnormal cases in real-world applications). As a result, researches in the past literature have focused on unsupervised learning approaches, which do not require any labeled data instances in the training set. One limitation of this line of approaches is that they often impose strong assumptions on the distribution of normal and abnormal instances, and any deviation from such assumption will cause the model to misbehave. The semi-supervised setting is a compromise between supervised and unsupervised approaches, which assumes that the training set consists of only normal instances. Many recent studies are dedicated to this line of approaches as it is often easy to collect clean data containing only normal cases. The research of this paper falls under the semi-supervised learning setting.

Given rapid advances in deep learning approaches in anomaly detection, we mainly focus on reviewing deep learning-related methods in this section.
\subsection{Deep Learning based Approaches}

Deep learning methods for time series anomaly detection can be generally classified into forecasting-based and reconstruction-based methods.

Forecasting-based models make predictions on the current data point given the previous points, and the anomaly is identified if the predicted value of the current data point deviates substantially from its ground truth value. All regression-based models for sequential tasks belong to this branch, while the individual difference between these models lies in their underlying model architectures that encode inherent patterns within the temporal sequences. RNN-based architectures (e.g., Long Short-Term Memory (LSTM)~\cite{graves2012long} and Gated Recurrent Unit (GRU)~\cite{chung2014empirical}) and CNN-based architectures (e.g., Temporal Convolution Neural Networks (TCNs)~\cite{oord2016wavenet}) are widely used to model data with sequential patterns. Despite their capabilities to capture temporal dependencies, these models are not designed to explicitly capture the inter-series relationship. For instance, LSTM or GRU projects the input features at a specific time step into a dense vector, while the pairwise relationship among time series is not explicitly learned. Whereas TCNs captures the temporal patterns for each time series separately and aggregate them through simple average or sum pooling.

Reconstruction-based models aim to learn the distribution of the normal data samples, by encoding them in a limited feature space through a reconstruction process, and a sample is identified as an anomaly if its reconstruction error is substantially larger than that of the normal samples. Representative works in this line of research are methods based on deep autoencoders (AEs)~\cite{su2019robust}~\cite{malhotra2016lstm}~\cite{park2018multimodal}, and Generative Adversarial Networks (GANs)~\cite{audibert2020usad}~\cite{li2019mad}. However, most reconstruction methods still adopt RNN-based or CNN-based neural network architecture to encode the temporal sequences, which limited their capability to model the relationship within multivariate time series.

\subsection{Capture Inter-dependencies through Graphs}
Recent developments of Graph Neural Networks (GNNs) has increased its adoption in time series anomaly detection to model the inter-series relationship. 

GDN~\cite{deng2021graph} treats each time series as a node with one-hot embedding as node features to construct a graph, which is encoded through GNNs to assist a downstream forecasting task. MTAD-GAT~\cite{zhao2020multivariate} learns a feature representation of each time series by aggregating information from similar time series or historical observations through an inter-series graph and an intra-series graph, which are constructed based on raw time series within a context window. The learned representation is used as input to both a reconstruction task and a forecasting task. Instead of using raw features or one-hot vectors as node features which may have a limited power to capture temporal patterns, FuSAGNet~\cite{han2022learning} constructs an inter-series adjacency matrix based on the latent feature representations generated from a sparse auto-encoder. Another recent work, GRELEN~\cite{zhang2022grelen} aims at reconstructing the time series through an VAE, where the latent variable is enforced to capture the inter-series relationship learned through graph structure learning. The aforementioned approaches generally assume a static feature-wise relationship globally or within the context window of a given sample, thus they cannot be used explicitly to detect deviation of inter-series relationship between normal and abnormal states.


Apart from the above methods, some works construct inter-series graphs dynamically across time and perform a graph forecasting or reconstruction task. For instance, EvoNet~\cite{hu2021time} extracts representative multivariate time series segments as nodes and learns a transition probability among them during training. However,  this model is designed to capture the temporal pattern shifts, not the inter-series relationship changes, between the segments at different periods. MSCRED~\cite{zhang2019deep}, a method similar to our design, aims at reconstructing a series of similarity matrices at different scales, which are encoded through an attention-based convLSTM~\cite{shi2015convolutional} network and decoded through CNN layers. However, convLSTM assumes the value of nearby pixels (of an adjacency matrix) are more related, which is actually not the case for graph-structured data. 
Meanwhile, the signature matrices are similarity matrices with the entry calculated based on point-wise cosine distance, which has limitations in measuring the un-synced sequences (e.g., lags exist between pairwise time series). Moreover, the training process is slow due to the reconstruction process of signature matrices at multiple scales. 


\section{Problem Formulation}
Following commonly used notations, the $i^{th}$ row and $j^{th}$ column of an arbitrary 2D matrix $A$ are denoted by $A_{i\cdot}$ and $A_{\cdot j}$ respectively. The $(i,j)^{th}$ entry of A is denoted by $A_{ij}$. The $l_2$-norm of a vector is represented by ${\| \cdot \|}_2$, while the Frobenius norm for a matrix is referred as $\|\cdot\|_F$.

In this study, We are given a multivariate time series $X \in R^{N\times T}$, collected over $T$ time ticks with $N$ features, as a training set. The goal is to detect anomalies in a test set, which is a time series with $N$ features collected over a different span of time ticks from the training set. We consider a semi-supervised setting, where the training set consists of only normal data and the test set contains a mixture of normal and anomalous data.

The input sample $S^t$ is a sub-sequence extracted from $X$ ($X$ is split into a series of sub-sequences with stride 1), which represents a collection of time ticks within a sliding context window of length $c$: $S^t=\{X_{\cdot i}\}_{i=t-c+1}^t$, where $X_{\cdot t} \in R^{N}$ is the multivariate time series at timestamp $t$. We aim at assigning an anomaly score to each time tick in the test set given a test sample, which is later threshold-ed to a binary label with $0$ being normal and $1$ being abnormal.

\section{Methodology}

\subsection{Overview}
\begin{figure*}[t!]
    \centering
        \includegraphics[width=1\linewidth]{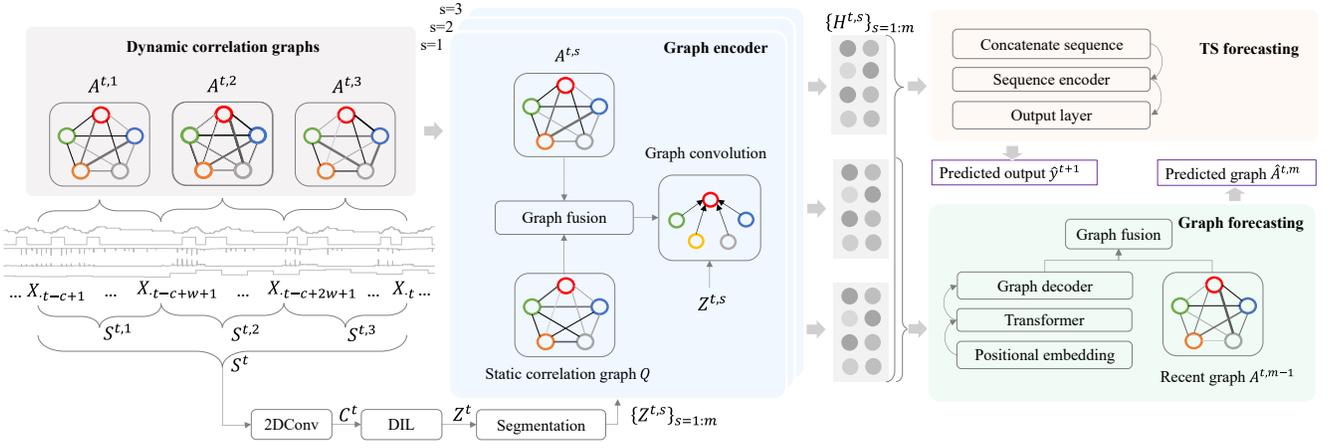}  

    \caption{The pipeline of DyGraphAD: A list of dynamic correlation graphs are constructed based on DTW at the beginning (take step size $m=3$, and a multivariate time series with $N=5$ (5 nodes) as an example). Next, the graph encoder takes the graphs as inputs and transforms them into a list of latent representations. Finally, the feature representations of the graphs are fed into the time series (TS) forecasting module and the graph forecasting module to obtain the one-step ahead predicted multivariate time series value and the predicted latest dynamic correlation graph.}
    \label{fig:framework}
\end{figure*}

In this section, we give a brief overview of the proposed framework: At the beginning, a series of inter-series graphs are constructed for each input sample in the pre-processing stage. Next, these graphs are encoded into hidden representations through a graph encoder module. The feature representations of graphs then serve as the input to a graph forecasting module and a time series forecasting module. The goal of the former is to predict the latest graph based on historical graphs, while the latter forecasts the time series value at this time step based on past values. The two tasks are optimized jointly during training, and generate a hybrid score at test time to detect anomalies. The overall pipeline is shown in Figure~\ref{fig:framework}.

\subsection{Construction of Dynamic Correlation Graph}
\label{section: similarity matrix}
Before training, each input sequence is broken into a list of disjoint segments in order to construct a series of similarity matrices. The purpose of building these matrices is to facilitate the graph forecasting task, i.e., given a set of historical similarity matrices, the goal is to predict the one-step ahead future matrix, which should be readily available during training.

Previous works usually construct the similarity matrix based on inner product or Euclidean distance for pairwise time series~\cite{choi2021deep}. However, those approaches are not suitable for pairwise time series that are out of sync, which is common in real-world scenarios. For instance, in industrial systems, one process may have cascading effects on the other processes, resulting in lags among multiple temporal sequences. 
Dynamic time warping distance (DTW)~\cite{sakoe1978dynamic} is a robust distance metric which is invariant to time shifts compared to the regular Euclidean distance metric. We adopt the fast C implementation of DTW from \cite{yurtman2021wannesm} which solves the optimal alignment match problem using a dynamic programming technique.

Formally, given each sample $S^t=\{X_{\cdot i}\}_{i=t-c+1}^t$, where $c=m \cdot w$, $m$ sequential adjacency matrices $\{A^{t,s}\}_{s=1}^{m}$ are constructed,  where $A^{t, s}\in R^{N\times N}$ is calculated based on the time series segment $S^{t,s} = \{X_{\cdot t-c+w(s-1)+1}, ..., X_{\cdot t-c+ws}\}$ with a window size of $w$ (see Figure~\ref{fig:framework} for a visual example). $w$ and $m$ are two hyper-parameters tuned for each specific dataset. In our implementation, $w$ and $m$ together determine the length $c$ of $S^t$.
An entry in one adjacency matrix represents the normalized similarity (i.e., between 0 and 1) between pairwise time series. Since DTW calculates an unbounded distance between two temporal sequences, we convert the entry of adjacency matrix $A^{t,s}$ within the range of 0 and 1 by
\begin{equation}
    A_{i,j}^{t,s}=\exp\left(-\frac{DTW^2(S_{i\cdot}^{t,s}, S_{j\cdot}^{t,s})}{\tau}\right)
\end{equation}
where $\tau$ is a hyper-parameter. $\{A^{t,s}\}_{s=1}^m$ is referred to as dynamic correlation graphs throughout this paper.

\subsection{Graph Encoder Module}
\label{section: graph feature extraction}
The graph encoder module encodes the dynamic correlation graphs into low-dimensional feature representations, with the purpose of capturing the temporal dynamics between graphs and the inherent structure within each graph. Specifically, an input sample is first broken into segments within a time window aligned with each dynamic correlation graph. Then, a graph convolution layer is applied on top of each graph with a latent representation of its corresponding segment as node features to obtain the hidden representations for each graph. 

Formally, given a multivariate time series sample $S^t \in R^{N \times c}$, we first feed it into a 2D convolution layer, with a kernel size $(1, 1)$ and a channel size of $d$, to obtain an initial representation $C^t \in R^{d \times N \times c}$. Then a dilated inception layer (DIL)~\cite{wu2020connecting}, with a kernel combination of $\{2, 3, 5, 7\}$ and a channel size of $d$, is applied to obtain a feature representation $Z^t \in R^{d \times N \times c}$.
The justification of the DIL layer is that aggregating nearby points through different kernel sizes can help the model to learn the similarity between pairwise time series up to a certain time shift, thereby mimicking the operation of DTW distance. Moreover, the DIL layer can later be stacked multiple times, with its dilation factor growing exponentially to capture long-range temporal patterns. 

Next, the feature representation $Z^{t}$ is broken into $m$ segments $\{Z^{t, s}\}_{s=1}^m$ along the time dimension coinciding with the same time window for each of the dynamic correlation graphs. Each of the time series segment $Z^{t,s} \in R^{d \times N \times w}$, together with an adjacency matrix capturing the inter-series relationship at step $s$, will be later used as the inputs to a graph convolution module to obtain a final feature representation of each graph.

Instead of directly using the dynamic correlation graph $ A^{t,s}$, we use a new matrix $\Tilde{A}^{t,s}$ as the adjacency matrix for the graph convolution module, which is calculated as a weighted combination of $A^{t,s}$ and a global static adjacency matrix $Q \in R^{N \times N}$
\begin{align}
    \label{eq. generate weighted graphs}
    & \Tilde{A}^{t,s} = \sigma(W_1) \odot Q + (1-\sigma(W_1)) \odot A^{t,s} \\
    & Q_{ij} = f_a (\xi_{i}\mathbin\Vert \xi_{j})
\end{align}
where $\xi_i\in R^{d}$ is a learnable embedding representing time series $i$, $W_1 \in R^{N \times N}$ is a learnable weight matrix and $\sigma$ is a sigmoid function.  $f_a$ is a graph generator function, i.e., a two-layer fully connected neural network with an output dimension $1$. $\odot$ is the element-wise multiplication and $\mathbin\Vert$ is the concatenation operation. 
Intuitively, the dynamic correlation graphs capture the short-term evolution patterns for the inter-series correlations at different steps, whereas the static correlation graph captures the long-term correlations, and impose constraints such that the learned node features are close to each other through the series of the graphs.
Thus combining them enables us to capture both effects.
Furthermore, the weighted matrix $\Tilde{A}^{t,s}$ improves the performance for both graph and time series forecasting tasks compared to using the DTW similarity matrix alone, which is validated in our experiments. 

Upon obtaining $\Tilde{A}^{t,s}$, we feed it together with $Z^{t,s}$ into a MixHop graph convolution~\cite{wu2020connecting} module in sequence to obtain a list of hidden states $\{H^{t,s}\}_{s=1}^m$, where $H^{t,s} \in R^{d \times N \times w}$ contains the node feature representations of each graph $\Tilde{A}^{t,s}$. 


\subsection{Graph Forecasting Module}
\label{section: graph forecasting}
Graph forecasting module, which consists of an encoder for a sequence of graphs, a decoder, and a recent graph update component, takes the list of graph hidden states $\{H^{t,s}\}_{s=1}^{m-1}$ from the graph encoder module as inputs and produces a single graph $\hat{A}^{t,m} \in R^{N \times N}$ as the output. 

The encoder consists of $L$ Transformer blocks on top of a learnable positional encoding layer. Specifically, we take the mean pooling along the time axis for each of the graph hidden states, and then concatenate the outputs into $P^t \in R^{N \times m \times d}$, i.e., $m$ hidden states with dimension $d$ for each of the time series. Then, the positional embedding and the Transformer blocks are applied along the second dimension, in order to preserve the positional information within the sequence of graphs, and learn the temporal dynamics among them. We adopt a causal attention mask for the multi-head attention mechanism of Transformer blocks to avoid information being leaked from future graphs to past graphs. Next, the aggregated hidden representation $O^t \in R^{N \times d}$ for the list of graphs is obtained by applying mean pooling along the segment (second) dimension of the outputs from the Transformer. Finally, the graph decoder converts the hidden state $O^t$ into an $N \times N$ adjacency matrix $E^t \in R^{N \times N}$
\begin{align}
    \label{eq. graph forecast: output pooling}
    & J^t = f_d(O^t) \\
    \label{eq. graph forecast: output decode}
    & E^t = (J^t){(J^t)}^T
\end{align}
where $f_d$ is a two-layer fully connected neural network with an output dimension $d$, followed by a $l_2$ normalization on the last dimension. 

Instead of directly adopting $E^t$,  we combine the most recent graph with it to obtain a predicted graph $\hat{A}^{t, m}$ at the $m^{th}$ step:
\begin{align}
    \label{eq. graph forecast: recent graph update}
    \hat{A}^{t, m} = \sigma(W_2)\odot E^t+(1-\sigma(W_2))\odot A^{t,m-1}
\end{align}
where $W_2 \in R^{N \times N}$ is a learnable weight matrix. The intuition is that the graph structure often varies smoothly over time rather than drastically under a relatively smaller time period, which makes the recent graph a good predictor for the future graph. Meanwhile, it compensates for the drawback of the graph encoder, which compresses the list of dynamic graphs into a sequence of low-dimensional states for efficient computation, but suffers from information loss under the data compression.

\subsection{Time Series Forecasting Module}
\label{section: time series forecasting}
One limitation of the graph forecasting task is its relatively low power to capture the temporal pattern within each time series, due to the discrete nature of the graphs by design: graphs are set apart from each other by a certain distance, resulting in a longer time span and better computational efficiency but a coarser version of the original time series.

The time series forecasting module is designed to mitigate the drawbacks of the graph forecasting task. The module couples the graph encoder with the temporal convolution layers to make prediction about the next-step ahead time series values. As demonstrated in other graph-based frameworks (e.g., GDN and MTAD-GAT), aggregating information from similar time series through graph neural networks improves the performance of time series forecasting.

Formally, the graph hidden states $\{H^{t,s}\}_{s=1}^{m}$ are concatenated along the time axis, followed by a batch normalization~\cite{ioffe2015batch}, to obtain $M^t \in R^{d \times N \times c}$. Unlike the raw sample $S^t$, each time series in $M^t$ contains the information propagated from other time series based on the inter-series relationship through the MixHop graph convolution. Finally, we obtain the predicted time series value vector $\hat{y}^{t+1}$ at time $(t+1)$ as
\begin{align}
    & \hat{y}^{t+1} = f_o(\text{pooling}( f_1(C^t) \mathbin\Vert f_2(Z^t)) \mathbin\Vert f_3(M^t))
\end{align}
where $f_1$, $f_2$, and $f_3$ are 2D convolution layers having a kernel size of $(1, c)$ and a channel size of $d$, and $f_o:R^{d\times N}\rightarrow R^N$ is an output layer, i.e., a two-layer fully connected neural network. $C^t$ and $Z^t$ are the initial feature representation and DIL output obtained from the graph encoder. The concatenation and the sum pooling are both applied on the time (third) dimension. Note that here we adopt a simple sequence encoder in our experiments which produces reasonably well performance. However, our framework can be easily extended to other advanced architectures, for instance, by applying multiple stacked DIL layers on top of $M^t$.

\subsection{Objective Function}
The training objective is to minimize the forecasting loss of both the next-step ahead time series values and the latest dynamic correlation graph at each time tick. The loss function is
\begin{align}
    L = \sum_{t\in T_{\text{train}}}{\frac{1}{N}\|y^{t+1}-\hat{y}^{t+1}\|_2^2 + \frac{1}{N^2} \| A^{t,m} - \hat{A}^{t,m} \|_F^2}
    \label{eq.loss function}
\end{align}
where $y^{t+1}$ and $A^{t,m}$ are the multivariate time series value at time step $(t+1)$ and the $m^{th}$ dynamic correlation graph, given the input multivariate time series sample $S^t$. 

\subsection{Anomaly Score}
The anomaly score at $t$ for each time series $i$ is calculated as
    \begin{align}
        & \text{Err}^t_i = \frac{1}{\frac{1}{\text{Err}^t_{i, \text{ts}}} +\frac{1}{\text{Err}^t_{i, \text{graph}}}}
    \end{align}
where $\text{Err}^t_{i, \text{ts}}$ is a squared error of the $i^{th}$ predicted time series value at $t$ and $\text{Err}^t_{i, \text{graph}}$ is a mean squared error at the $i^{th}$ row 
 of the predicted dynamic correlation graph at t. We adopt a harmonic mean of both errors as the final score for each time series since the two scores have different scales. Finally, the anomaly score at $t$ is obtained as the average score of $N$ time series.
 


\section{Experiment}

\begin{table*}[t!]
\small
    \setlength{\abovecaptionskip}{0cm}
    \centering
        \caption{Performance comparison (F1 (\%)) on real-world datasets, precision (Pre (\%)) and recall (Rec (\%)) are reported for reference. The best performance is highlighted in bold. The results of baseline models are partially from~\protect\cite{su2019robust},~\protect\cite{audibert2020usad} and~\protect\cite{zhao2020multivariate}.}
    \setlength{\tabcolsep}{1.65mm}{
    \begin{tabular}{ccccccccccccccccc}\\ \toprule
               &   \multicolumn{3}{c}{\textbf{SWaT}} & \multicolumn{3}{c}{\textbf{WADI}} & \multicolumn{3}{c}{\textbf{SMAP}} & \multicolumn{3}{c}{\textbf{MSL}}& \multicolumn{3}{c}{\textbf{SMD}} \\ \midrule
               \textbf{Methods} & \textbf{F1} &  \textbf{Pre} & \textbf{Rec} & \textbf{F1} &  \textbf{Pre} & \textbf{Rec} & \textbf{F1} &  \textbf{Pre} & \textbf{Rec}  & \textbf{F1} &  \textbf{Pre} & \textbf{Rec}   & \textbf{F1} &  \textbf{Pre} & \textbf{Rec}  \\ \midrule
   Deep SVDD   & 82.87 & 97.85 & 71.86 & 67.33 & 67.06 & 67.60 & 69.67& 67.68 &71.77 & 85.01 &89.87 & 80.65 & 86.30 & 86.58 & 87.49  \\
   DAGMM       &81.83 & 92.02 & 73.68 & 44.34 & 56.92 & 36.31 & 71.05 & 58.45 & 90.58 & 70.07 & 54.12 & \textbf{99.34} & 70.94 & 59.51 & 88.82 \\
   LSTM-VAE    & 84.95 & \textbf{99.73} & 73.99 & 67.52 & 86.52 & 55.37 & 72.74 & 57.65 & \textbf{98.53} & 90.58 & 88.48 & 92.78 & 78.42 & 79.22 & 70.75 \\ 
   MAD-GAN     & 86.53 & 90.85 & 82.60 & 70.51 & 76.22 & 65.60 & 88.14 & 94.22 & 82.79 & 91.38 & 85.55 & 98.07 & 47.85 & 45.96 & 64.25   \\ 
   USAD        & 84.60 & 98.70 & 74.02 & 57.25 & 72.39 & 47.34 & 86.34 & 76.97 & 98.31 & 92.72 & 88.10 & 97.86 & 94.63 & 93.14 & 96.17   \\ 
   OMNI& 85.28  & 99.44 & 74.65  & 41.02 & 64.86 & 29.99 & 84.34&74.16 & 97.76 & 89.89 & 88.67 & 91.17 & 88.57 & 83.34 & 94.49\\ 
   MSCRED       & 84.59 &92.69  & 77.80    & 55.01& 51.16  &59.48& 69.33 & 94.11 & 54.88  & 82.05 & 87.37 & 77.34   & 78.75 & 84.73 &79.11   \\ 
   MTAD-GAT    & 85.50 & 98.24 & 75.69 & 67.94 & \textbf{87.90} & 55.37 & 90.13 & 89.06 & 91.23 & 90.84 & 87.54 & 94.40 & 92.72 &91.50 & 95.49  \\ 
   GDN         & 89.57 & 96.14 & 83.84 & 71.54 & 72.92 & 70.21 &92.31 & 91.64 & 92.98 & 94.62 & 94.91 & 94.34 & 94.84 & 94.08 & 96.16 \\ \midrule
   DyGraphAD    &\textbf{92.31} &94.29	& \textbf{90.42}  & \textbf{85.38}   & 84.13 & \textbf{86.66} & \textbf{94.94}	&93.39& 96.55 & \textbf{95.92}	& \textbf{97.01}	& 94.86 & \textbf{96.55}	&\textbf{95.14}	& 98.19 \\      
   wo. TS & 89.57	& 92.62	& 86.72 & 83.80 & 81.13 & 86.66 &87.49 &\textbf{95.93} & 80.41 & 93.34 & 91.87& 94.86 & 94.20 &	92.36 &	96.42 \\
   wo. Graph & 87.18	& 91.55	& 83.20 & 77.05 & 72.41 & 82.32 & 91.33 & 87.01 & 96.10 &  94.43 & 93.44 & 95.43 & 96.11 & 93.80 & \textbf{98.71} \\

   \bottomrule
    \end{tabular}
    }
    \label{tab:comparison_result}
\end{table*}

\subsection{Datasets}

In this study, five real-world benchmark datasets are used for the evaluation of time series anomaly detection methods: SWaT~\cite{mathur2016swat}. WADI~\cite{ahmed2017wadi},  SMD~\cite{su2019robust},  SMAP and MSL~\cite{hundman2018detecting}.  SWaT and WADI are released by iTrust Center~\footnote{\url{https://itrust.sutd.edu.sg/itrust-labs_datasets/dataset_info/}} to support research in cybersecurity. SMD is a server machine dataset collected from a large internet company. It contains 28 entities, with each being trained separately. SMAP and MSL datasets are collected from a spacecraft of NASA.
Every dataset has a training set containing only normal data samples, and a separate test set containing a mixture of normal and abnormal samples. Similar to \cite{deng2021graph}, we down-sample the original data of SWaT and WADI to 1 measurement every 10 seconds by taking the median value. Additionally, datasets are normalized by min-max scaler before the training. 

\subsection{Baseline Methods and Evaluation Metrics}
 The baseline methods used in our experiments are Deep SVDD~\cite{ruff2018deep}, DAGMM~\cite{zong2018deep}, LSTM-VAE~\cite{park2018multimodal}, MAD-GAN~\cite{li2019mad}, USAD~\cite{audibert2020usad}, OMNI-ANOMALY~\cite{su2019robust}, MSCRED~\cite{zhang2019deep}, MTAD-GAT~\cite{zhao2020multivariate} and GDN~\cite{deng2021graph}. Certain methods, especially forecasting-based approaches, can achieve relatively better performance when their anomaly scores are normalized. Therefore, we apply the same scaling function in~\cite{deng2021graph}, which standardizes the anomaly scores using the median and interquartile range (IQR) of the training set, on all the baseline methods. We report the best result between the non-scaled and the scaled version. However, our model does not require any scaling on the anomaly scores, as the performance based on non-scaled scores is similar to the one based on scaled scores.

F1-score is used as a performance metric to evaluate the performance of anomaly detection, which is defined as
\begin{align}
    & F1 = 2\cdot \frac{P\cdot R}{P+R} \\
    & P = \frac{TP}{TP+FP} \\
    & R = \frac{TP}{TP+FN}
\end{align}
where P is the precision, R is the recall, and TP, FP, TN, FN stand for true positive, false positive, true negative, and false negative respectively. The threshold for all models is chosen as the one which achieves the best F1 over the test sets. In addition, we follow the point-adjusting strategy suggested in \cite{su2019robust}, which labels a whole anomaly segment as 1 as long as one of the points within the segment is detected as an anomaly.

\subsection{Experimental Setup}
Our model is trained with an Adam optimizer~\cite{kingma2014adam} at a learning rate of 0.001 for 10 epochs. $20\%$ of the training set is selected as a validation set where the best epoch is stored based on the validation loss. The hidden dimension is set to $d=64$, and the number of transformer blocks is set to $L=2$. The temperature $\tau$ used to construct the dynamic graphs for SWaT, SMAP, WADI, MSL, SMD are 1, 1, 5, 5, 0.5, respectively (we restrict the choices of this parameter to a limited set, $\{0.1, 0.5, 1, 5, 10\}$). The step size and the window size are set to $m=6$ and $w=5$ for all of the datasets.

\subsection{Comparison Study}
Table~\ref{tab:comparison_result} shows the results of comparison of all models: DyGraphAD has a better balance between precision and recall, which can be verified from the high F1 values compared to the baseline methods among all datasets (The precision and recall of DyGraphAD are on average among the top 3). In comparison, the other graph-based approaches, such as GDN and MTAD-GAT, also achieve relatively better detection performance than the majority of baseline methods. These methods leverage the inter-series relationship to assist a time series forecasting task, while anomalies are detected via one-step ahead forecasting accuracy. However, DyGraphAD still outperforms both of them in terms of F1, demonstrating a distinct advantage of explicitly utilizing dynamically evolving inter-series relationships for anomaly detection. In addition, most methods perform badly in WADI dataset due to a low recall rate. Specifically, WADI has a much lower anomaly rate compared to other datasets, making the anomaly events difficult to be captured. However, our model achieves a significant improvement in WADI compared to all baseline methods. One reason is that WADI consists of a large number of sensors, making the inter-series relationship more complex and vital to anomaly detection.

We also provide the performance of the time series task and the graph forecasting task that is trained separately in the last two rows. While each task is able to achieve a reasonably well performance, combining both further improves the detection accuracy. Meanwhile, optimizing both tasks together is more computationally efficient compared to training each of them separately and then combining the scores (the training time for joint optimization is almost equivalent to training each of the tasks separately). Meanwhile, the optimization goal of one task can be beneficial to the other, i.e., both of them focus on learning a better feature representation of the inter-series relationship, which is helpful in predicting both the latest graph and next-step ahead time series value.

\subsection{Ablation Study}
To evaluate the effects of each sub-modules on our framework, we perform an ablation study on SWaT, WADI, and SMAP by excluding components in the default model. The results are summarized in Table~\ref{tab:ablation_result}. For the graph forecasting task, we test three scenarios: (1) removing the state update process, i.e., by only using the most recent graph to predict the current graph.  (2) removing the recent graph. (3) removing the static graph (from the graph encoder) and the recent graph. The results agree with our assumptions that the forecasting performance depends on the short-term evolving dynamics of the historical graphs as well as the long-term relationship, and combining the recent graph directly can further improve the performance. For the time series forecasting module, two settings are explored: (1) removing the static graph. (2) removing the static graph and dynamic graph, i.e., by removing the graph convolution entirely. All the settings achieve relatively lower F1 than the default setting, implying both the short-term and long-term inter-series relationships are beneficial for the forecasting task.

\begin{table}[h!]
\small
    \setlength{\abovecaptionskip}{0cm}
    \centering
    \caption{Performance comparison (F1 (\%)) when core components of our model are removed.} 
    \begin{tabular}{ccccc}\\ \toprule
     \textbf{Settings}          &   \textbf{SWaT} & \textbf{WADI} & \textbf{SMAP}  \\ \midrule
   DyGraphAD &\textbf{92.31} &\textbf{85.38} & \textbf{94.94} \\ \midrule
   DyGraphAD(Graph)   & 89.57 & 83.80 & 87.49  \\
   (Graph) recent graph only       & 87.31   & 69.49 & 85.10 \\  
   (Graph)  wo. recent graph       & 88.58   &  78.19  & 87.04 \\ 
   (Graph)  wo. recent\&static graph    & 86.90  & 71.82 & 86.59    \\ \midrule
   DyGraphAD(TS)                       &87.18 & 77.05 & 91.33   \\
   (TS) wo. static graph       &  85.53 &  74.50 & 89.69 \\ 
   (TS) wo. static\&dynamic graph   & 84.43  &  65.01 & 89.05 \\ 
   \bottomrule
    \end{tabular}

    \label{tab:ablation_result}
\end{table}

\begin{figure*}[t!]
\centering
\begin{subfigure}[t]{0.49\textwidth}
    \centering
              \includegraphics[height=4.2cm]{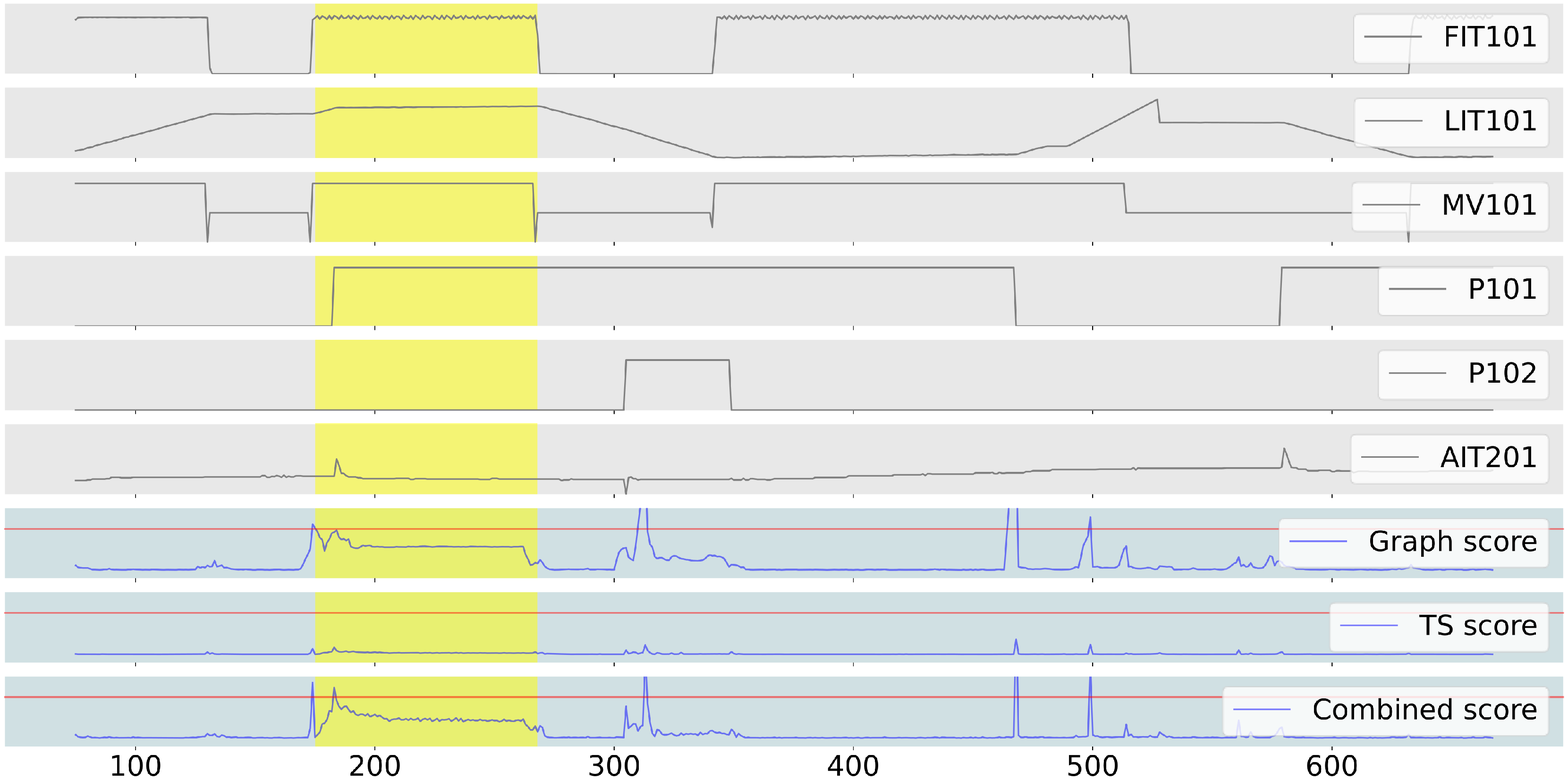}
    \caption{}
    \label{fig: swat_ts_unidentified}
\end{subfigure}
\hspace{\fill} 
\centering
\begin{subfigure}[t]{0.49\textwidth}
    \centering
           \includegraphics[height=4.2cm]{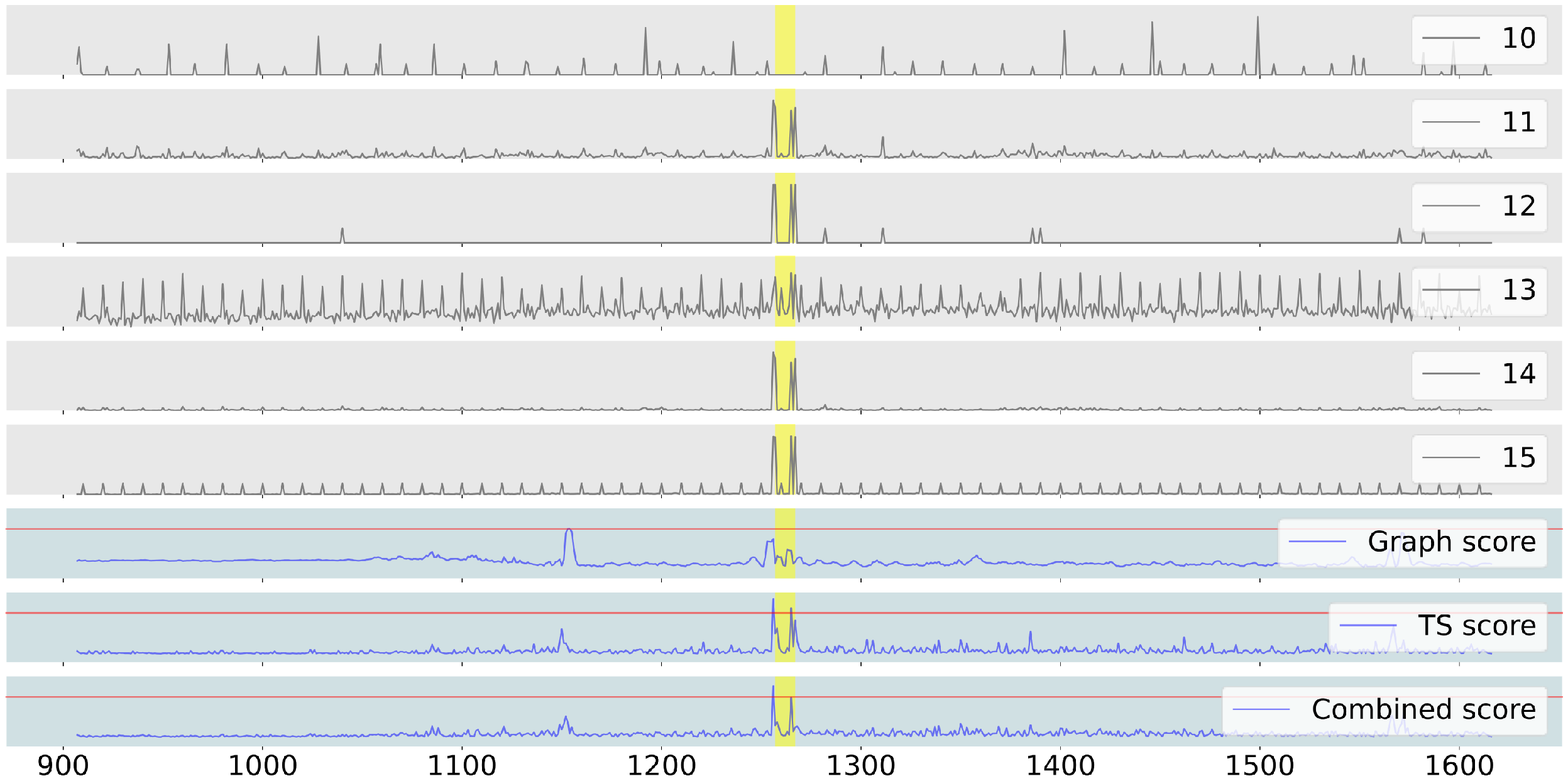}
    \caption{}
    \label{fig: smd_graph_unidentified}
\end{subfigure}
\centering
\begin{subfigure}[t]{0.49\textwidth}
    \centering
       \includegraphics[height=4.2cm]{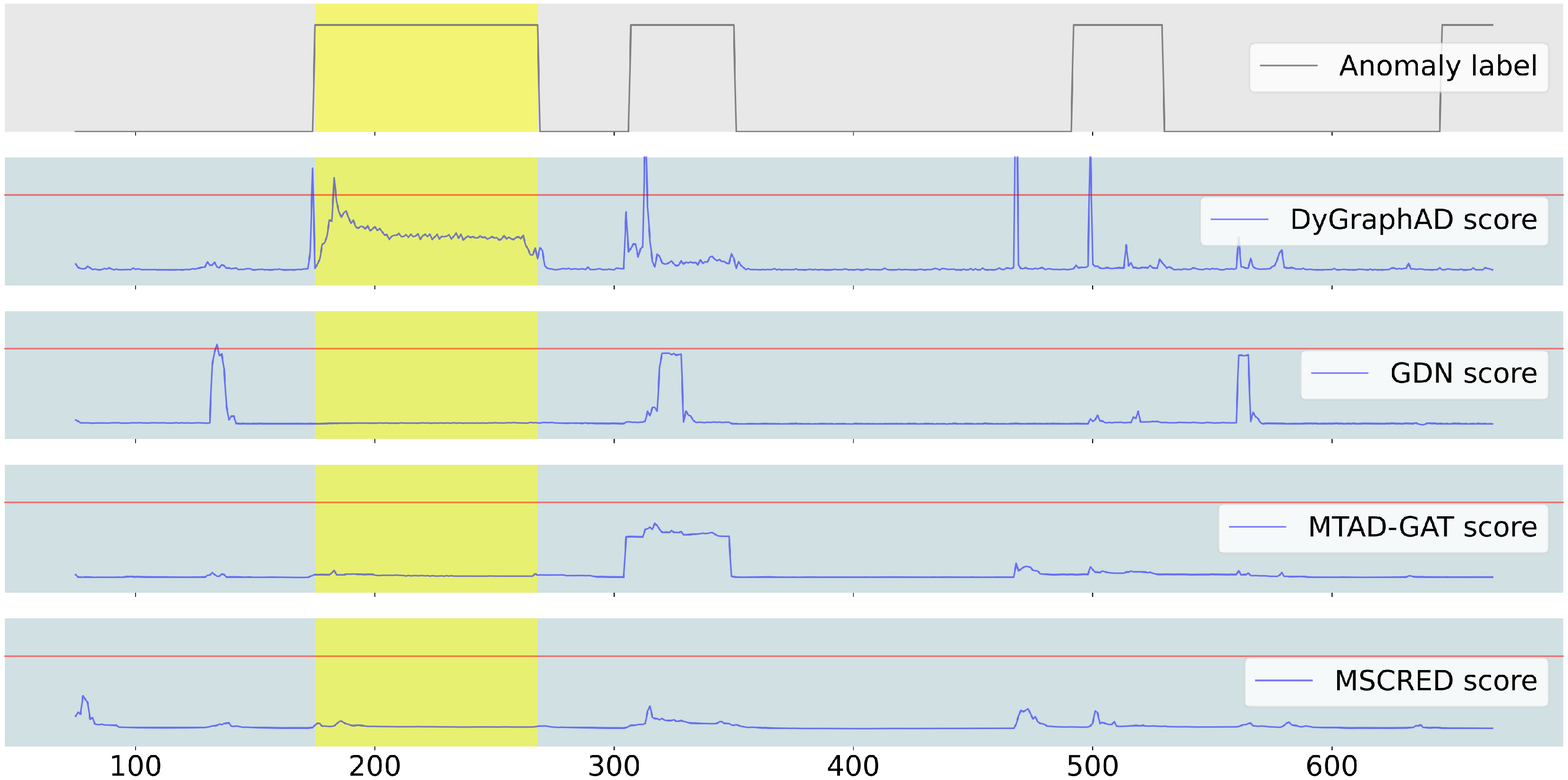}
    \caption{}
    \label{fig: swat_event-1_case_study}
\end{subfigure}\hspace{\fill} 
    \centering
 \begin{subfigure}[t]{0.49\textwidth}
 \centering
   \includegraphics[height=4.2cm, width=8.3cm]{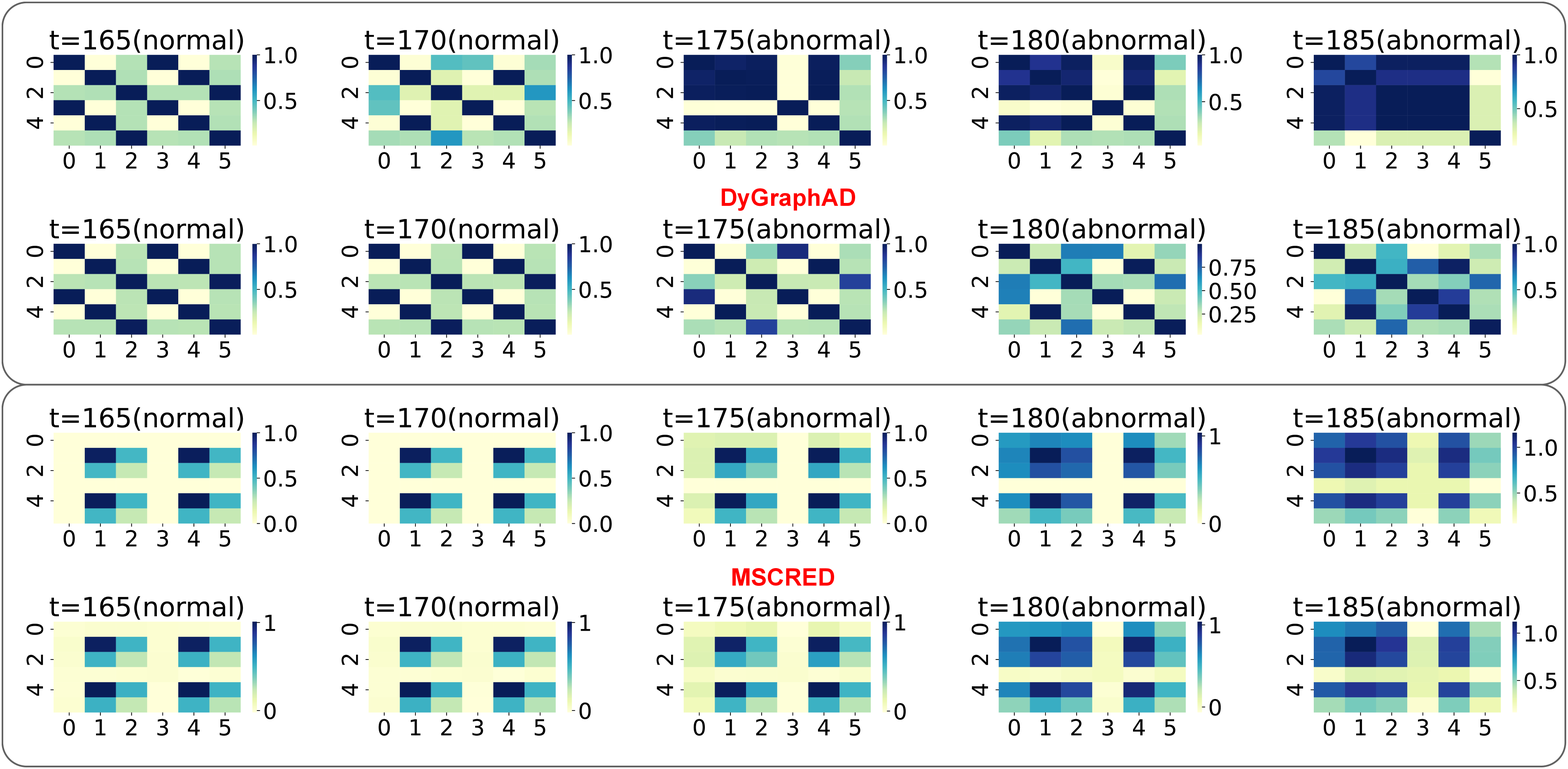}  
    \caption{}
    \label{fig: swat_event-1_adjs_mscred}
\end{subfigure}
\caption{(a) \& (b) are time series segments from the SWaT and SMD datasets; 
anomaly scores and corresponding thresholds from the graph forecasting task (Graph score), the time series forecasting task (TS score), and the combination of them (Combined score) are shown in the bottom three rows, 
 respectively. Regions highlighted in yellow are anomalies that are only captured by either the graph forecasting task or the time series forecasting task. (c) is the anomaly scores from the graph-based models corresponding to the time series segment in (a). (d) is the dynamic correlation graphs constructed around the highlighted anomaly in (a). The top two rows in (d) are the ground truth and predicted graphs of DyGraphAD, while the bottom two rows are the ground truth and reconstructed graphs of MSCRED. DyGraphAD generates large error scores for graphs during the abnormal period, while MSCRED misses the anomaly by reconstructing the graphs well in both the normal and abnormal regions.}

\end{figure*}

\subsection{Case Study}

\subsubsection{Capturing Anomalous Graph Evolution Patterns}

\begin{figure}[H]
\centering
\begin{subfigure}[t]{0.9\linewidth}
    \centering
               \includegraphics[height=2.9cm]{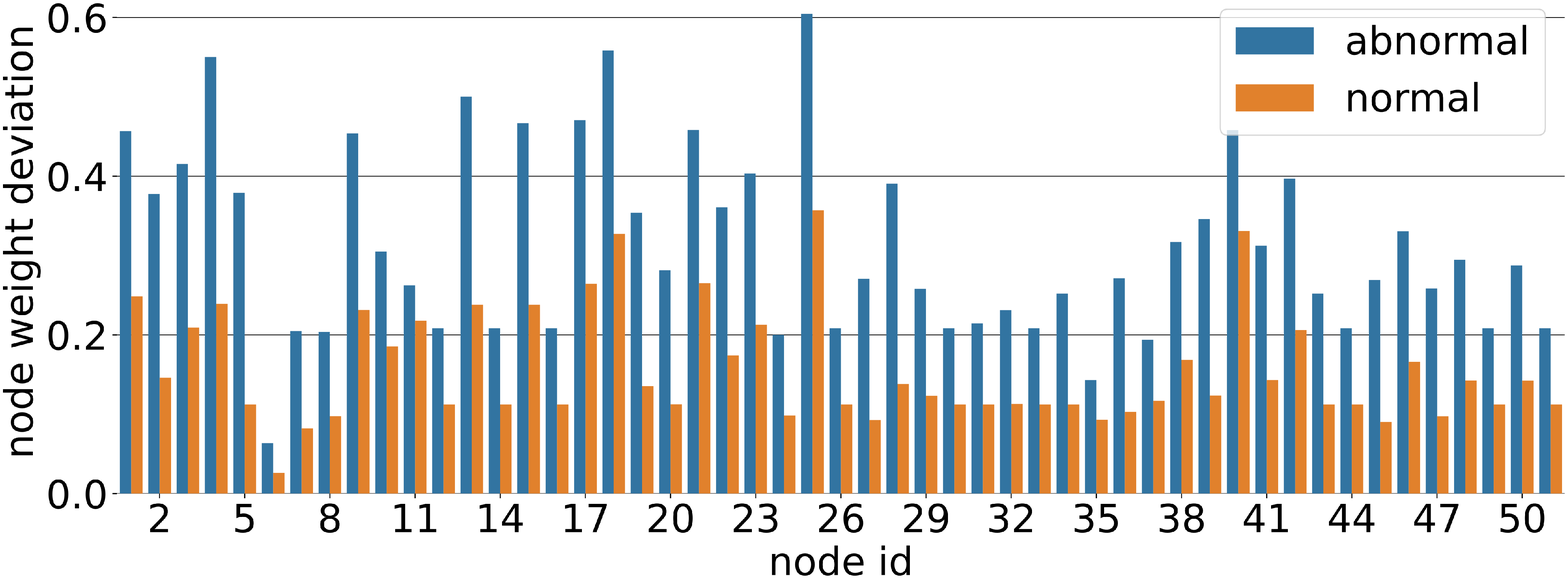}
    \caption{SWaT (51 features/nodes)}
    \label{fig: swat graph deviation}
\end{subfigure}
\newline
\centering
\begin{subfigure}[t]{0.9\linewidth}
    \centering
          \includegraphics[height=2.9cm]{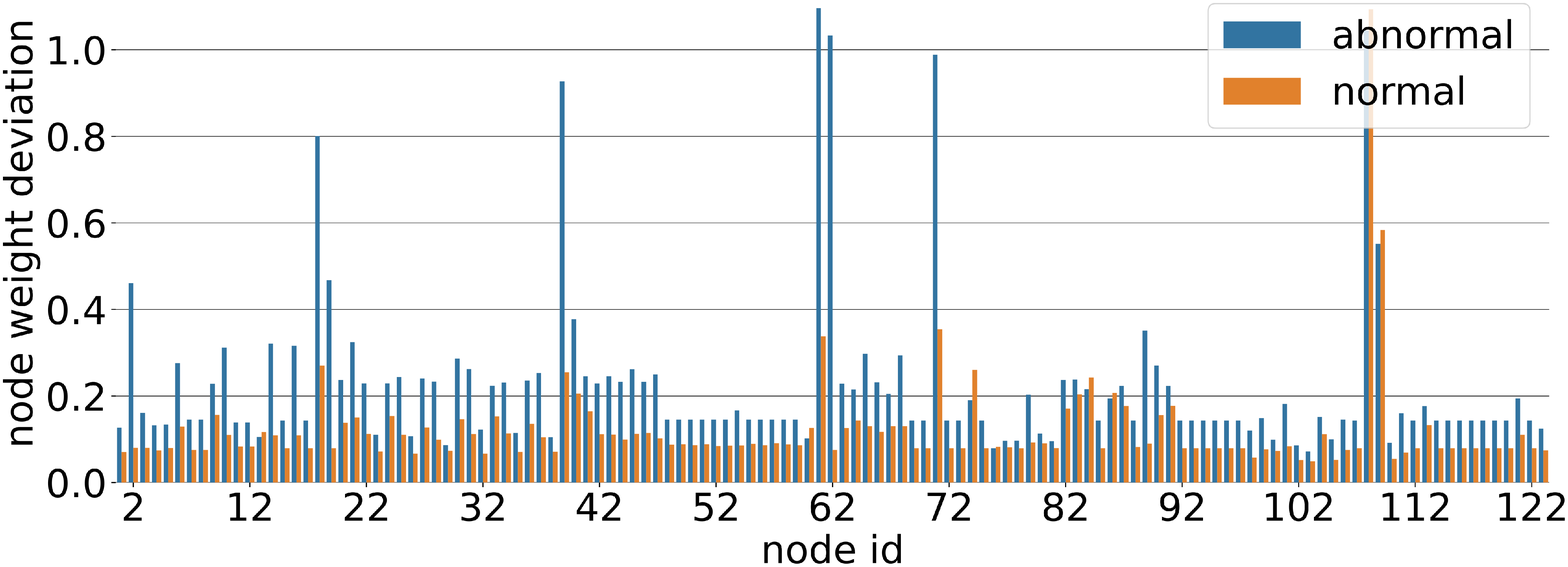}
    \caption{WADI (123 features/nodes)}
    \label{fig: wadi graph deviation}
\end{subfigure}
\caption{Deviation of node weight (sum of edge weights per node) versus feature/node id for SWaT and WADI datasets in normal states (normal), or during a transition from a normal state to an abnormal state (abnormal). On average, the node weight during a transition into an abnormal state varies more drastically than into a normal state.}
\label{fig: graph deviation}
\end{figure}

The basic assumption for our framework is that the graph evolution pattern in the normal state differs from that of the abnormal state. We demonstrate this idea by plotting the deviation of the node weight (sum of edge weights per node) between adjacent graphs in Figure~\ref{fig: graph deviation}: For both the SWaT and WADI datasets, the weight deviation per node of the dynamic correlation graphs at normal states is much higher compared to the case during a transition from a normal state to an abnormal state. This agrees with our assumption that the inter-series relationship at an abnormal state tends to be different from the past history, and our model is able to leverage this information in order to improve detection performance. In fact, even using the recent graph alone to predict current graph can produce reasonably well performance (as shown in Table~\ref{tab:ablation_result}).

\subsubsection{Combined Effects of Time Series and Graph Forecasting}
We examine the combined effects of the time series and graph forecasting tasks through visual inspections in Figure~\ref{fig: swat_ts_unidentified}\ \&~\ref{fig: smd_graph_unidentified}.  First, the graph forecasting task is more sensitive to the feature-wise information, we demonstrate this idea through an example from the SWaT dataset: The highlighted anomaly shown in Figure~\ref{fig: swat_ts_unidentified}) is caused by a cyber-attack. Specifically, when the water treatment process operates normally, water can only be pumped into the tank when the tank level is not full. Therefore, the water flow, which is measured by the sensor ``FIT101'', can be high only after the tank level, measured by ``LIT101'', goes down. However, the motorized valve ``MV101'', which controls the water flow into the tank, is forced open during a cyber-attack in the yellow region. Consequently, the attack causes tank overflow, resulting in a high value for both ``LIT101'' and ``FIT101''. As shown in the bottom part of Figure~\ref{fig: swat_ts_unidentified}, the graph forecasting task is able to identify this anomaly via an abnormal feature-wise relationship. Although this anomalous event can be possibly identified through an abnormal seasonal pattern, the time series forecasting error score around this anomaly is much smaller compared to the other anomalies (the trapezoid shape highlighted in yellow is similar to the historical patterns, resulting in a smaller forecasting error). In fact, there are many similar anomalies in SWaT and WADI datasets, where the temporal patterns around these anomalies do not deviate much from the historical sequences. Therefore, the time series forecasting task generally performs worse than the graph forecasting task in these datasets (as shown in Table~\ref{tab:comparison_result}). 

On the other hand, the short anomaly from SMD dataset in Figure~\ref{fig: smd_graph_unidentified} is easily identified by the time series forecasting task, but fails to be captured by the graph forecasting task alone. This is because the temporal dynamics within the graphs are in a coarser granularity than the actual time series, resulting out of ignorance of certain anomalies with very short spans. There are many other similar anomalies in the SMD, MSL and SMAP datasets, which can be detected via temporal patterns in a finer granularity. As a result, the time series forecasting task has a better performance than the graph forecasting task in those datasets (as shown in Table~\ref{tab:comparison_result}).

In essence, although both tasks are able to learn temporal patterns and feature-wise relationships to some extent, each of them has a better focus on a different data dimension. Thus, combining them can eventually improve the anomaly detection performance (the global score is able to capture those anomalies as shown in the bottom section of Figure~\ref{fig: swat_ts_unidentified} and Figure~\ref{fig: smd_graph_unidentified}). 

\subsubsection{Comparing against Other Graph-Based Methods}
\label{section: comparison study for swat data}
To further explore the differences between different graph-based methods, we plot their anomaly scores corresponding to the time series segment in Figure~\ref{fig: swat_ts_unidentified}. 
Figure~\ref{fig: swat_event-1_case_study} shows that all of the models, except DyGraphAD, fail to capture the highlighted anomaly since their error scores are below the red threshold line. In contrast, DyGraphAD is able to identify this anomaly based on a large deviation between the predicted graph evolving pattern and the ground truth pattern (shown in Figure~\ref{fig: swat_event-1_adjs_mscred}). On the other hand, MSCRED, an approach that also identifies anomalies based on graph deviations, has generated a score substantially lower than its threshold. The bottom two rows of Figure~\ref{fig: swat_event-1_adjs_mscred} provide an explanation: Specifically, MSCRED accurately reconstructs its similarity matrices in both the normal and abnormal phases. This phenomenon traces back to the issue associated with the autoencoder, which tends to learn an identity mapping function if its underlying model architecture does not have the right degree of compression to extract the salient features from the normal samples. In addition, MSCRED adopts a graph reconstruction process based on convLSTM layers as encoder and deconvolution layers as decoder, which has limitations in modeling the graph-structured data where adjacent points are not necessarily related. In contrast, our model circumvents the aforementioned issues by adopting a forecasting framework with a graph encoder that is more suitable for graph-structured data.


\section{Conclusion}
In this work, we introduce a multivariate time series anomaly detection framework, DyGraphAD, which detects anomalies based on both the deviation of inter-series relationship and intra-series patterns from abnormal states to normal states. Our experiment results show that incorporating the dynamically evolving inter-series relationship into the model design, as well as combining a graph forecasting task with a time series forecasting task, can substantially improve the performance of anomaly detection. The competitive results on real-world industrial datasets demonstrate the superiority of our framework against a variety of anomaly detection techniques. 
\appendix





\bibliographystyle{plain}
\bibliography{reference}


\end{document}